\documentclass[runningheads]{llncs}
\usepackage{caption}
\usepackage[T1]{fontenc}
\usepackage{float}
\usepackage{graphicx}
\usepackage[hyphens]{url}
\usepackage[hidelinks]{hyperref}
\usepackage{makecell}
\usepackage{listings}
\usepackage{xcolor}
\begin{document}

\title{Structured Four-Stage Legal Translation:} 
\subtitle{From Natural-Language Traffic Rules to PROLOG}
%
\author{May Myo Zin \inst{1}\orcidID{0000-0003-1315-7704} 
\and Wachara Fungwacharakorn \inst{1}\orcidID{0000-0001-9294-3118} 
\and Ken Satoh \inst{1}\orcidID{0000-0002-9309-4602} 
\and Katsumi Nitta \inst{1}\orcidID{0000-0002-9018-8603}}
\authorrunning{M.M. Zin et al.}
%
\institute{Center for Juris-Informatics, ROIS-DS, Tokyo, Japan \email{\{maymyozin,wacharaf,ksatoh\}@nii.ac.jp, knitta00@gmail.com}}

\maketitle              
\begin{abstract}

Traffic regulations are written for human interpretation and therefore rely on shared background knowledge and flexible phrasing, which inherently introduce ambiguity, context dependence, and semantic underspecification. These linguistic characteristics conflict with the precision required by computational reasoning engines such as Prolog, which demand explicit logical structure. This study evaluates two baseline translation approaches, Natural Language to Prolog ($NL\rightarrow Prolog$) and Logical English to Prolog ($LE\rightarrow Prolog$), and introduces a new reasoning-guided translation framework called Structured Four-Stage Legal Translation ($S4L\rightarrow Prolog$). The proposed S4L framework performs semantic role extraction, scene completion, logical mapping, and Prolog rule generation within a single guided prompt, enabling direct translation of raw traffic rules into executable logic without human intervention. A benchmark consisting of twenty real-world traffic rules was used to evaluate each approach in terms of syntactic validity, semantic correctness, and logical completeness. $S4L\rightarrow Prolog$ achieves the highest accuracy, correctly formalizing 75 percent of the rules, while $NL\rightarrow Prolog$ reaches 60 percent and $LE\rightarrow Prolog$ reaches 55 percent. Qualitative analysis further shows that S4L captures implicit causal relations, deontic modality, and exception structure more reliably than the baselines. These results demonstrate that structured reasoning prompts can substantially improve the reliability of natural-language-to-logic translation for legal and safety-critical applications.

\end{abstract}

\section{Introduction}
Autonomous vehicles and intelligent transport systems must obey complex traffic regulations originally drafted for human understanding. These rules describe obligations, prohibitions, and permissions that govern safe and lawful conduct on public roads. However, the semantic clarity required by machines differs fundamentally from that intended for humans. Natural-language rules are ambiguous, context-dependent, and often underspecified, relying on shared background knowledge about road configurations, visibility, or intent. Translating these rules into formal logic suitable for computational reasoning therefore remains a major challenge in legal informatics and autonomous systems research.

Existing efforts to mechanize the translation of natural-language norms into formal logic remain limited in their ability to achieve full automation. Controlled Natural Languages (CNLs), such as Attempto Controlled English (ACE) \cite{fuchs2008attempto} and Logical English (LE) \cite{kowalski2023logical}, reinterpret rules in a human-readable yet logically precise form, enabling their subsequent mapping to formal logic. While these methods enhance clarity and reduce ambiguity, they still rely heavily on manual paraphrasing and the explicit inclusion of implicit contextual information by domain experts. Automating the transformation of natural language into CNLs is also a challenging task. Formal logic systems for traffic and safety compliance, including those based on deontic, temporal, and defeasible logic, provide rigorous reasoning once the formal rules are established. However, they assume the existence of a formalized rule base, which is both costly and reliant on human effort. While deep learning and large language model (LLM) approaches exhibit fluent linguistic interpretation, they often generate syntactically incorrect or semantically incomplete logic in formal translation tasks, unless the input is first simplified or restructured through human intervention. Despite advances across these paradigms, there remains no fully automated workflow that starts from raw natural-language traffic rules, performs semantic interpretation, world-knowledge inference, and normative reasoning, and produces executable logic ready for deployment. 

This paper presents a structured four-stage prompt that guides an LLM through a complete interpretive cycle, enabling the translation of natural language traffic norms into executable Prolog logic. The goal is to achieve logical completeness and semantic fidelity without requiring human intervention for contextual supplementation. Rather than relying on manual preprocessing, as in CNL pipelines or hybrid approaches combining pseudo-CNL with LLMs, or on shallow rule extraction typical of deep learning and NLP techniques, our system operates directly on raw natural-language traffic rules. 

\section{Related Work}

Efforts to translate natural-language regulations into machine-interpretable form span two converging threads: (i) domain-specific formalization of traffic rules and (ii) NLP-to-logic pipelines for extracting and encoding normative content.

In the context of traffic regulation, early work manually-formalized subsets of national and international traffic laws into temporal logics to enable automated compliance checking, thereby demonstrating the value of temporal operators for rules that unfold over time. A representative example is Maierhofer et al.’s formalization of German interstate rules into metric temporal logic (MTL) with a four-stage process (extract, concretize, define predicates, and synthesize formulas), which remains a touchstone for structuring end-to-end pipelines even when NLP is introduced later \cite{maierhofer2020formalization}. Rizaldi et al. \cite{rizaldi2015formalising,rizaldi2017formalising}, followed by Linker et al. \cite{linker2017spatial}, encoded subsets of the Vienna Convention on Road Traffic in Isabelle/HOL, focusing on rule compliance and collision avoidance. These works demonstrated the feasibility of mechanically proving safety properties but relied heavily on manual formalization by domain experts. Complementary efforts employed spatio-temporal logics such as Multi-Lane Spatial Logic (MLSL) and Signal Temporal Logic (STL) to capture notions like safe distance and right-of-way \cite{hilscher2011abstract,esterle2020formalizing}. In parallel, frameworks such as Responsibility-Sensitive Safety (RSS) \cite{shalev2017formal} and Rulebooks \cite{censi2019liability} provided prioritized sets of constraints guiding safe and interpretable motion planning. While these approaches improved automation in checking rule compliance, they still required manual translation of textual norms into logical formulas. 

A parallel line of work targets NLP-assisted translation to temporal and deontic logics. Manas and Paschke \cite{manas2023semantic} propose an SRL-assisted pipeline that identifies predicate–argument structures and temporal aspects from rule text, then maps them to temporal logic. Building on this, TR2MTL \cite{manas2024tr2mtl} uses LLMs with chain-of-thought prompting to translate traffic rules into MTL in a human-in-the-loop setting, reporting strong accuracy on a curated traffic-rule dataset. Recent engineering surveys echo this need to (a) translate free-text traffic rules, (b) check semantic correctness, and (c) include expert review due to linguistic ambiguity \cite{wan2024semantic}.

Beyond temporal logics, legal-tech standards aim to capture normative structure explicitly. LegalRuleML \cite{palmirani2011legalruleml} provides an XML-based representation tailored to legal norms and has accompanying transformations to (defeasible) deontic logic for automated reasoning; these works emphasize traceability of deontic effects (obligation, permission, prohibition) and interoperability with rule engines.

Complementary studies examine automatic extraction of legal norms and evaluate NLP tools for identifying obligations, prohibitions, conditions, and exceptions in statutory text—capabilities that underpin upstream information extraction in NLP-to-logic pipelines \cite{ferraro2019automatic}. Most existing frameworks adopt a hybrid, human-in-the-loop paradigm \cite{zin2025towards}. Natural language processing components such as semantic role labeling, dependency parsing, and large language model prompting generate structured candidate representations. These candidates are then encoded by formal reasoning back-ends, after which domain experts validate semantics and resolve ambiguities before deployment. Recent research has refined this interaction loop through techniques such as grammar-constrained decoding and verification-guided prompting, aiming to improve the faithfulness and reliability of the generated formal expressions \cite{englishgrammar}.

\section{Research Gap and Motivation}

Despite the substantial body of work reviewed above, a key gap persists at the intersection of linguistic interpretation and logical execution. Existing pipelines rely on human pre-processing to resolve ambiguity and add contextual detail before formalization \cite{zin2025towards}. LLMs, on the other hand, show emerging capacity for contextual understanding but often conflate meaning with syntax when asked to generate logic directly.

The motivation for the present study stems from the need to integrate interpretive reasoning and logical formalization within a single automated process. Traffic regulation offers an ideal testbed because its rules combine concrete spatial conditions with abstract normative modalities. Successful automation in this domain would demonstrate how LLMs can bridge the gap between natural semantics and machine-readable logic. 

Accordingly, this research pursues three objectives:
\begin{itemize}
    \item To design a structured prompt that compels the LLM to perform stepwise semantic, contextual, and logical reasoning.
    \item To evaluate whether such scaffolding yields syntactically valid and semantically faithful Prolog representations without human intervention.
    \item To empirically measure the framework’s effectiveness through a benchmark of 20 authentic traffic rules, assessing logical validity, contextual completeness, and semantic fidelity. 
\end{itemize}

By addressing these goals, the study aims to advance automated normative translation from descriptive linguistic statements to executable reasoning artifacts, which is a foundational step toward explainable and legally compliant autonomous systems.

\section{Methodology}

This study investigates three distinct strategies for translating natural-language traffic regulations into executable Prolog rules. Two approaches, Logical English to Prolog ($LE\rightarrow Prolog$) and Natural Language to Prolog ($NL\rightarrow Prolog$), serve as baseline models. The third approach, Structured Four-Stage Legal Translation ($S4L\rightarrow Prolog$), represents the proposed framework. 

\subsection{Baseline Methods}
\subsubsection{Logical English to Prolog ($LE\rightarrow Prolog$):}
The first baseline employs Logical English (LE)\footnote{The Logical English (LE) inputs used in this study follow a pseudo-Logical English format: a semi-controlled variant of English that mirrors the syntactic and modal structure of formal Logical English but does not strictly conform to its grammar specification.
This pseudo-LE representation facilitates interpretability by large language models while maintaining semantic alignment with deontic expressions such as \textit{permitted}, \textit{prohibited}, and \textit{obligatory}.}, a controlled natural language designed to reduce ambiguity while maintaining human readability.
The model receives a 19-shot prompt, containing nineteen examples of LE rules paired with their Prolog translations. This method evaluates the LLM’s capacity for syntactic transformation under structured, low-ambiguity input conditions, where most semantic interpretation is already encoded in the LE syntax.

\subsubsection{Natural Language to Prolog ($NL\rightarrow Prolog$):}
The second baseline uses the unaltered natural-language form of each rule as input. The same 19-shot prompt structure is used, but examples consist of ordinary traffic rules written in natural language and their corresponding Prolog forms. This baseline measures the LLM’s ability to perform direct language-to-logic mapping, handling implicit meanings, contextual dependencies, and deontic expressions without prior formalization.

\subsection{Proposed Method: Structured Four-Stage Legal Translation}

The proposed \textit{\textbf{S}tructured \textbf{Four}-Stage \textbf{L}egal Translation} ($S4L$) framework is founded on the observation that linguistic comprehension and logical representation are not separate stages of cognition but complementary layers of understanding. A human expert, when reading a traffic rule such as “Do not cross a solid white line”, implicitly reconstructs a scene containing two or more lanes, recognizes that the solid line demarcates a boundary that should not be crossed, and classifies the rule as a prohibition applying to a driver or vehicle. The LLM is guided to emulate precisely this interpretive sequence, moving systematically from semantic understanding to formal codification. This is achieved by breaking down the translation task into four consecutive reasoning phases: semantic role extraction, scene completion, logical mapping, and formal output generation. Each phase is explicitly represented in the prompt. Unlike the baselines, $S4L\rightarrow Prolog$ is a zero-shot prompt, relying solely on a single detailed instruction template rather than in-context examples. The complete process is composed of four reasoning stages that together operationalize linguistic interpretation into executable PROLOG logic. 

\subsubsection{Stage 1 - Semantic Role Extraction:}

In the first stage of processing, the model analyzes the surface structure of a traffic rule to extract its underlying semantic components. This involves breaking down the rule into parts that describe \textit{who is doing what, under what conditions, and whether the action is allowed or not}. 
These components include:

\begin{itemize}
    \item[] \textit{\textbf{Agent}} – the entity responsible for the action
    \item[] \textit{\textbf{Primary Action}} – the main behavior being regulated
    \item[] \textit{\textbf{Method Action}} – the manner or means by which the primary action is performed 
    \item[] \textit{\textbf{Condition}} – the explicit or implicit circumstances under which the rule applies
    \item[] \textit{\textbf{Modality}} – the type of normative force involved: obligation, prohibition, or permission 
    \item[] \textit{\textbf{Exception}} – any explicitly stated exceptions to the rule (if present)
    \item[] \textit{\textbf{Implicit Facts}} – background knowledge or assumptions necessary for understanding the rule, even though they are not explicitly stated
\end{itemize}

\noindent This process extends conventional semantic role labeling and frame semantics by incorporating \textit{deontic modalities}, which express duties, permissions, or prohibitions. To clarify, we present the following example:

\begin{example}
\label{ex:solid-line}
\textit{“If a driver is driving on a road with a solid white line, he must not use the oncoming lane when overtaking.”}
\end{example}

\noindent
From this traffic rule, the model automatically extracts the following semantic roles:

\begin{itemize}
    \item \textit{\textbf{Agent:} driver}
    \item \textit{\textbf{Primary Action:} overtake}
    \item \textit{\textbf{Method Action:} use the oncoming lane}
    \item \textit{\textbf{Condition:} driving on a road with a solid white line}
    \item \textit{\textbf{Modality:} prohibition}
    \item \textit{\textbf{Exception:} none}
\end{itemize}

\noindent
In addition, it automatically infers several \textit{\textbf{implicit facts}} essential for a complete understanding of the rule:

\begin{itemize}
    \item \textit{Roads with a solid white line have at least two lanes, one for each direction of traffic.}
    \item \textit{A solid white line separates opposing flows of traffic.}
    \item \textit{“Using the oncoming lane” means entering the lane intended for traffic in the opposite direction.}
    \item \textit{Overtaking is normally done by changing lanes.}
    \item \textit{The solid white line prohibits crossing.}
\end{itemize}

\noindent
By explicitly identifying both semantic roles and implicit facts, the framework transforms complex natural language into structured, machine-interpretable data.

\subsubsection{Stage 2 - Scene Completion:}
The second stage addresses a crucial shortcoming of previous translation methods: the absence of world knowledge. Most traffic laws presuppose an elaborate context, such as the presence of lanes, signage, opposing directions of traffic, and physical boundaries. However, these contextual elements are rarely stated explicitly. The scene completion step instructs the model to reconstruct this context in concise natural language before formalization.

For example, the overtaking rule (as introduced in Example~\ref{ex:solid-line}) leads to the following scene description:
“\textit{A driver is traveling on a road divided by a solid white line, which separates two lanes of opposing traffic. The driver considers overtaking a slower vehicle ahead. To overtake, the driver would need to cross the solid white line and enter the oncoming lane. However, because of the line marking, crossing into the oncoming lane for overtaking is prohibited in this situation.}”.

This contextual reconstruction serves a dual function. First, it validates the model’s understanding of the rule. Second, it connects linguistic meaning to the physical situation the rule refers to. By requiring the model to explicitly describe the scene, the framework ensures that each logical rule is grounded in a consistent and accurate understanding of the physical environment.

\subsubsection{Stage 3 - Logical Mapping and Short Explanation:}
The third stage transforms the enriched natural-language understanding into formal logic. Here, the model converts the semantic and contextual information into PROLOG predicates consistent with a traffic reasoning ontology. Each predicate corresponds to an entity or relation inferred in the earlier stages, while deontic modality is encoded through meta-predicates such as \textit{obligation/1}, \textit{prohibition/1}, or \textit{permission/1}.

For instance, the earlier overtaking rule (as introduced in Example~\ref{ex:solid-line}) yields the following logical clause:

\begin{lstlisting}[language=Prolog, basicstyle=\scriptsize, linewidth=\textwidth]
prohibited(overtake(Driver, Vehicle)) :-
    driving_on(Driver, Lane1),
    driving_on(Vehicle, Lane1),
    adjacent(Lane1, Lane2),
    oncoming_lane(Lane2, Lane1),
    separated_by_line(Lane1, Line, Lane2),
    solid_white(Line),
    do_by(overtake(Driver, Vehicle), use_lane(Driver, Lane2)).
\end{lstlisting}

As part of this stage, the model is instructed to generate a short explanatory paragraph that links the natural language rule to its formal representation, as follows: \\

\noindent \textbf{\textit{Short Explanation:}}\\
\textit{This rule prohibits overtaking by using the oncoming lane when the two lanes are separated by a solid white line. The do\_by/2 predicate captures the fact that the prohibition applies to overtaking by using the oncoming lane — not overtaking in general. Implicit facts about lane adjacency, directional flow, and road markings are used to complete the logic. This ensures the rule is only enforced in contexts where crossing the line would be illegal.}\\ [-8pt]

An additional instruction is given to the model to maintain predicate consistency by reusing predicates from a predefined list when appropriate. This mechanism helps preserve alignment with existing ontology terms and ensures semantic consistency across generated representations. However, the handling of newly introduced predicates is not addressed in the current implementation. In future work, this aspect will be explored through a semi-automated review process that combines LLM-based predicate similarity checking with human validation. The design and evaluation of this process are beyond the scope of the present paper.

\subsubsection{Stage 4 - Formal Output Generation:}
In the final stage, the model outputs the results in a standardized structure containing the original rule, the extracted semantic roles, the scene description, the executable Prolog code, and a short explanatory paragraph linking the natural language and formal representation. This consistency supports both human review and downstream automation. 

This format also serves a pedagogical function: it provides an interpretable audit trail showing how the LLM derived each logical component from the original rule. In legal and safety-critical contexts, such traceability is vital for accountability and verification.

\section{Experimental Setup}

To support reproducibility, we consolidate here all experimental conditions used across the three translation methods. The two baselines ($LE\rightarrow Prolog$ and $NL\rightarrow Prolog$) used a 19-shot prompt containing paired examples of rules and translations, whereas $S4L\rightarrow Prolog$ used a single zero-shot prompt specifying the four-stage translation structure. All experiments were run using the GPT-4.1 model with deterministic decoding (temperature = 0) to eliminate randomness. A reference inventory of 60 predefined predicates (derived from existing manual Prolog representations of traffic rules) is provided to all translation methods.  The inventory contains Prolog predicates that capture common relational and unary concepts used in reasoning about traffic rules (e.g., \texttt{driving\_on/2}, \texttt{overtake/2}, \texttt{separated\_by\_line/3}, \texttt{prohibited/1}), and it serves as a semantic reference to encourage consistency across translations. However, the models are not strictly constrained to this set: when a suitable predicate is unavailable or semantically inadequate, the model is allowed to introduce new predicate names following Prolog syntax and relational logic conventions.

\subsection{Prompt Templates}

As illustrated in Listings~\ref{lst:NLLEProlog} and~\ref{lst:S4LPrompt}, the prompt templates define the structural and procedural foundations for the translation process. Listing~\ref{lst:NLLEProlog} specifies the format for $NL/LE\rightarrow Prolog$ conversion, while Listing~\ref{lst:S4LPrompt} outlines the four-stage reasoning structure adopted in $S4L\rightarrow Prolog$.
\\

\begin{lstlisting}[language=Prolog, caption={Prompt template for $NL/LE\rightarrow Prolog$ translation.}, basicstyle=\scriptsize, linewidth=\textwidth, label={lst:NLLEProlog}]
You are an assistant tasked with translating traffic rules into PROLOG code.

Using the provided pre-defined predicates and examples as a guide, write the PROLOG code for the given traffic rule. Do not provide explanations.

If there are predicates with similar meanings 
(e.g., driving_on/2 is the same as travelling_on/2, use_lane/2, use_road/2, etc.), 
do not create new predicates unless necessary.

Existing predicates: [pre-defined predicate list omitted for brevity]

Example 1:
Original Traffic Rule (NL/LE version): 
             <<originalTrafficRule_NL[1]>> or <<originalTrafficRule_LE[1]>>
Prolog Code: <<prolog[1]>>
...
...
...
Example 19:
Original Traffic Rule (NL/LE version): 
             <<originalTrafficRule_NL[19]>> or <<originalTrafficRule_LE[19]>>
Prolog Code: <<prolog[19]>>

Given Original Traffic Rule (NL/LE version): 
             <<originalTrafficRule_NL>> or <<originalTrafficRule_LE>>
Prolog Code:

\end{lstlisting}

\begin{lstlisting}[caption={Brief structure of the $S4L\rightarrow Prolog$ prompt. The complete prompt is available at \href{https://github.com/mtproleg/NLL2FR2025/blob/main/S4L-Prolog-Prompt.pdf}{the GitHub link}.},label={lst:s4l-brief}, basicstyle=\scriptsize, linewidth=\textwidth, label={lst:S4LPrompt}]
System role: Expert in computational linguistics, deontic logic, and autonomous driving reasoning.

Task: Translate natural-language traffic rules into executable Prolog logic for an autonomous reasoning engine.

Four-stage reasoning structure:
  STEP 1 - Semantic Role Extraction
  STEP 2 - Scene Completion
  STEP 3 - Logical Mapping
  STEP 4 - Output Formatting

Output includes:
  - Semantic roles, modality, exception, and implicit facts
  - Completed driving scene description
  - Executable Prolog representation
  - Short explanation of how the explicit and implicit meanings were combined
\end{lstlisting}

\subsection{Traffic Rules Dataset}

The dataset consists of twenty \textit{implicit traffic rules} derived from German court decisions interpreting the Road Traffic Act (\textit{Straßenverkehrsordnung}, StVO) with specific reference to \textit{Sign 295}, the solid white line road marking.  Each rule represents a judicially inferred interpretation that clarifies the legal meaning and permissible behavior associated with this traffic sign. The original rules were written in German and later translated into English for the purpose of this research. The translated rules were kept in their original legal phrasing to preserve authentic ambiguity and open-texture expressions typical of legislative drafting. This choice intentionally reflects real-world interpretive difficulty, rather than sanitized examples. The diversity of rule types allowed for systematic testing of the model’s ability to infer missing context, detect modality, and generate logically coherent formal representations.

Table~\ref{tab:traffic_rules} presents the complete list of traffic rules used for evaluation. The dataset covers a range of logical constructs including prohibitions, permissions, obligations, and conditional clauses.  
This diversity enables evaluation of how well each method captures deontic modality, causal reasoning, and spatial relationships within the formal Prolog translation task.

\vspace{-10pt}
\begin{table}[h!]
\scriptsize
\centering
\caption{Traffic Rules Dataset (Rules 1--20)}%
\label{tab:traffic_rules}
\begin{tabular}{p{0.8cm}p{11.1cm}}
\\
\Xhline{0.5pt}
\rule{0pt}{3ex}\textbf{Rule} & \textbf{Description} \\[2pt]
\Xhline{0.5pt}
\rule{0pt}{3ex}1 & If a driver is driving on a road with a solid white line, he must not use the oncoming lane when overtaking. \\

2 & At an intersection, if a driver is travelling in a lane designated for traffic travelling straight ahead or turning left and there is a lane to the driver's right designated for traffic travelling straight ahead or turning left and if the two lanes are separated by a solid white line, the driver must not move into the right lane by crossing the solid white line. \\

3 & If a driver is travelling on a carriageway with a solid white line, the driver must not cross the line to overtake a vehicle in front of him wishing to turn left and stopping in the carriageway because of oncoming traffic. \\

4 & If a driver is driving on a carriageway with a solid white line, he must not cross or drive on this line, neither for the purpose of turning left. \\

5 & If a driver is driving on a lane with a solid white line and the lane is so narrow that overtaking is not possible without crossing the solid white line, then he must not initiate an overtaking manoeuvre in the first place. \\

6 & If a driver is driving on a road with a solid white line, he may overtake if he does not touch the solid white line and, in accordance with §5 para. 4 sentence 2 StVO, keeps a sufficient distance to third party road users. \\

7 & If a driver is driving in a lane separated from other lanes by solid white lines, then the driver must not change lanes. \\

8 & If a driver is driving on a carriageway with a solid white centre line, then neither the body nor the cargo of the vehicle may protrude above the centre line. \\

9 & If a driver is driving on a carriageway with a solid white line, then a properly stored cargo may protrude above the solid white line if otherwise traffic to the right of the vehicle would be endangered, but oncoming traffic would not. \\

10 & If the driver of a vehicle is driving on a carriageway with a solid white line, then, under section 127 of the Criminal Procedure Code, he must not cross it in order to identify an offender having committed an offence without any harmful consequences. \\

11 & If a driver is driving on a carriageway with a solid white line, then he must not stop on the carriageway if he thereby obstructs other traffic or if the distance from the solid white line is smaller than 3 m. \\

12 & If a driver is travelling on a carriageway with a solid white line demarcating the carriageway to the right of a special path, then the driver may stop to the left of the solid white line. \\

13 & If a driver is driving on a carriageway with a solid white line demarcating the edge of the carriageway and if there is sufficient empty space to the right of the carriageway, parking or stopping to the left of the carriageway demarcation is not permitted. \\

14 & If a driver is driving on a carriageway with a solid white line demarcating the edge of the carriageway (edge line), he or she is allowed to drive across it. \\

15 & When driving on a road with a solid white line, the driver may cross it only in exceptional cases. \\

16 & If, at the beginning of a solid white lane, an overtaking manoeuvre has not yet been completed, the driver shall discontinue the overtaking manoeuvre if it can only be continued by using the other lane. \\

17 & If a driver drives on a carriageway with a solid white line and crosses this line, this does not constitute a violation of the prohibition to overtake in §5 of the Road Traffic Act (StVO). \\

18 & If a driver is driving on a carriageway with a solid white line, then he must not turn left into a property driveway if this is only possible by crossing the solid white line. \\

19 & If a driver is driving on a carriageway with a solid white edge line at the left-hand edge, he must not cross it to park on the verge behind. \\

20 & If a pedestrian crosses a carriageway the lanes of which are separated by a solid white line, then he may trust that no vehicle is coming from the left on the oncoming lane. \\[12pt]

\Xhline{0.5pt}
\end{tabular}
\end{table}

\section{Results and Discussion}


Each of the twenty benchmark traffic rules was independently translated using three methods: \textbf{$LE\rightarrow Prolog$}, \textbf{$NL\rightarrow Prolog$}, and the proposed \textbf{$S4L\rightarrow Prolog$}. All generated Prolog programs were manually assessed for three criteria: \textit{syntactic validity}, \textit{semantic correctness}, and \textit{logical completeness}.  A translation was marked as \textit{Correct} if the resulting Prolog clauses were executable and semantically faithful to the original traffic rule.  Outputs that were only partially correct or omitted causal relationships were marked as \textit{Incorrect}, while translations that demonstrated superior contextual or logical completeness were labeled as \textit{Best}. 

The complete outputs from all approaches, along with the corresponding human expert evaluations, are available in the online supplementary materials\footnote{\url{https://github.com/mtproleg/NLL2FR2025}}.  Table~\ref{tab:evaluation_results} summarizes the comparative results across all twenty rules. The overall counts are as follows: 

\noindent\textbf{Correct count:} $LE\rightarrow Prolog$ = 11, $NL\rightarrow Prolog$ = 12, $S4L\rightarrow Prolog$ = 15 

\noindent\textbf{Incorrect count:} $LE\rightarrow Prolog$ = 9, $NL\rightarrow Prolog$ = 8, $S4L\rightarrow Prolog$ = 5  

\noindent\textbf{Best outcomes:} $S4L\rightarrow Prolog$ = 3 (Rules 1, 4, 16).
\vspace{-10pt}
\begin{table}[h!]
\scriptsize
\centering
\setlength{\tabcolsep}{12pt} 
\caption{Comparative Evaluation of Translation Accuracy (Rules 1--20)}
\label{tab:evaluation_results}
\begin{tabular}{cccc}
\\
\Xhline{0.5pt}
\rule{0pt}{3ex}\textbf{Rule} & \textbf{\boldmath $LE\rightarrow Prolog$} & \textbf{\boldmath $NL\rightarrow Prolog$} & \textbf{\boldmath $S4L\rightarrow Prolog$} \\[2pt] \Xhline{0.5pt}
\rule{0pt}{3ex}1 & Correct & Correct & \textbf{Best} \\
2 & Correct & Incorrect & Correct \\
3 & Incorrect & Correct & Correct \\
4 & Correct & Correct & \textbf{Best} \\
5 & Incorrect & Incorrect & Incorrect \\
6 & Incorrect & Correct & Correct \\
7 & Correct & Correct & Correct \\
8 & Correct & Correct & Correct \\
9 & Correct & Incorrect & Correct \\
10 & Correct & Correct & Correct \\
11 & Correct & Incorrect & Incorrect \\
12 & Correct & Correct & Correct \\
13 & Incorrect & Incorrect & Correct \\
14 & Correct & Correct & Correct \\
15 & Incorrect & Correct & Correct \\
16 & Incorrect & Correct & \textbf{Best} \\
17 & Incorrect & Incorrect & Incorrect \\
18 & Incorrect & Incorrect & Correct \\
19 & Correct & Correct & Incorrect \\
20 & Incorrect & Incorrect & Incorrect \\[2pt] \Xhline{0.5pt}
\end{tabular}
\end{table}

\subsection{Quantitative Summary}

Across all twenty rules, \textbf{$S4L\rightarrow Prolog$} achieved the highest correct rate at approximately \textbf{75\%}, followed by \textbf{$NL\rightarrow Prolog$} at \textbf{60\%} and \textbf{$LE\rightarrow Prolog$} at \textbf{55\%}. Although the $LE\rightarrow Prolog$ method benefited from its syntactic structure, it often lacked contextual inference.  The $NL\rightarrow Prolog$ baseline captured natural semantics more effectively but misinterpreted logical operators in complex sentences. In contrast, $S4L\rightarrow Prolog$ consistently produced logically valid and semantically complete outputs, even under a zero-shot setting.

\subsection{Qualitative Analysis}

Several representative rules illustrate the comparative strengths and weaknesses of each approach:

\begin{itemize}
    \item[] \textbf{Rule 2 (Conjunction vs. Disjunction):}  $NL\rightarrow Prolog$ incorrectly used \textit{and(
    [straight\_ahead, turn\_left])} instead of \textit{or([straight\_ahead, turn\_left])}, while 
$S4L\rightarrow Prolog$ correctly inferred the disjunctive structure from context.
    
    \item[] \textbf{Rule 5 (Causal Relationship):}  All methods failed to represent the causal dependency between the narrow road condition and the prohibition of overtaking, indicating the need for enhanced world knowledge integration.
    
    \item[] \textbf{Rule 10 (Exception Priority):}  $S4L\rightarrow Prolog$ accurately modeled the interaction between the main rule and the exception clause, preserving the correct rule priority in its logical form.
    
    \item[] \textbf{Rule 16 (Ongoing Action):}  Only $S4L\rightarrow Prolog$ explicitly represented that overtaking was already in progress, demonstrating its superior ability to capture temporal context.
\end{itemize}

These examples highlight S4L’s capacity to manage deontic reasoning, relational predicates, and conditionally dependent clauses more effectively than the baselines.

\subsection{Discussion}

The comparison shows that structured reasoning guidance improves the reliability of LLM-generated logic translations. While the few-shot baselines performed adequately on syntactically simple rules, they often struggled with implicit conditions or exception handling. 
In contrast, S4L’s zero-shot framework effectively decomposed interpretation into interpretable cognitive stages, achieving superior logical integrity without relying on example-based prompting. S4L’s four-stage process (semantic role extraction, scene completion, logical mapping, and output formatting) consistently promoted internal consistency across predicates and modalities. The findings suggest that structured prompting can outperform example-based learning in legal text formalization tasks, particularly where implicit facts and deontic structure are essential. $S4L\rightarrow Prolog$ therefore provides a promising framework for transforming natural-language regulations into machine-actionable logic suitable for use in autonomous vehicle reasoning and other computational law applications.

\section{Challenges and Insights}
The translation of natural-language legal text into executable logic exposes the fundamental tension between human interpretive flexibility and machine formal rigidity. During experimentation, several challenges emerged that highlight both the promise and the boundaries of LLM-based legal formalization.

The first major challenge concerns linguistic ambiguity and vagueness. Legal drafters intentionally employ open-textured terms, such as \textit{reasonable distance}, \textit{safe manner}, or \textit{due care}, to preserve interpretive flexibility across contexts. While LLMs can paraphrase or substitute synonyms, they cannot by themselves assign concrete thresholds or numerical parameters to these concepts without external knowledge bases or policy directives. In our study, the model often handled such phrases descriptively, generating predicates like \textit{safe\_distance(Vehicle, FrontVehicle)} without quantifying what constitutes “\textit{safe}”. Although this maintains semantic fidelity, it limits the code’s executability in operational systems. The finding underscores that full automation of legal formalization ultimately requires integration with domain ontologies and empirical parameters. A second challenge concerns temporal reasoning. Many traffic norms are inherently temporal: they involve states that change over time, such as traffic lights, vehicle motion, or right-of-way at dynamic intersections. The underlying Prolog framework, while suitable for static deontic relations, lacks native temporal operators. Consequently, when the model translated rules like “\textit{Stop until the light turns green}”, it tended to produce static prohibitions rather than temporal conditions. Addressing this requires extending the logical target language to include temporal-deontic operators or coupling it with event calculus frameworks. A third insight involves exception hierarchy and conflict resolution. Legal norms rarely exist in isolation; they interact through priority structures such as emergency rules overriding ordinary ones. Although the LLM-generated rules correctly captured many “unless” exceptions, they did not consistently establish explicit precedence among conflicting norms. For autonomous reasoning engines, resolving such conflicts is critical. Incorporating non-monotonic logic or defeasible reasoning layers atop the generated Prolog code could provide a structured method for such resolution.

Finally, there is a conceptual insight concerning human–machine interpretive complementarity. Rather than replacing human legal reasoning, automation supports a synergistic division of labor: the LLM excels at enumerating plausible interpretations and reconstructing context rapidly, while human experts remain essential for validating normative soundness. The structured prompting approach thus acts as a bridge, capturing the interpretive richness of human reasoning and the formal rigor of computational logic.

\section{Conclusion and Future Work}

This study demonstrates that structured four-stage prompting (S4L) can transform large language models into semantic–deontic translators, capable of producing executable Prolog logic directly from natural-language traffic rules. The framework achieved high logical validity and interpretive completeness without human intervention. Beyond its empirical performance, an important contribution of the S4L approach is its transparent and auditable reasoning trail, which captures how the model interprets each rule before producing formal logic. This traceability is essential for legal and safety-critical applications such as autonomous driving, where explainability and accountability are essential. 

Future research will extend this work by integrating temporal-deontic operators, expanding ontology constraints, and employing reinforcement-guided refinement using expert feedback. Broader evaluation across multilingual legal corpora will test robustness and generality. Ultimately, this approach supports the development of legally interpretable autonomous systems, bridging the divide between normative text and computational reasoning.

\begin{credits}
\subsubsection{\ackname} This research was supported by the “Strategic Research Projects” grant from ROIS (Research Organization of Information and Systems), the “R\&D Hub Aimed at Ensuring Transparency and Reliability of Generative AI Models” project of the Ministry of
Education, Culture, Sports, Science and Technology, and JSPS KAKENHI
Grant Numbers JP22H00543, JP25H00522,  JP25H01112, and JP25H01152.

\end{credits}


\bibliographystyle{splncs04}
\bibliography{mybib}

@inproceedings{manas2023semantic,
  title={Semantic role assisted natural language rule formalization for intelligent vehicle},
  author={Manas, Kumar and Paschke, Adrian},
  booktitle={International Joint Conference on Rules and Reasoning},
  pages={175--189},
  year={2023},
  organization={Springer}
}

@inproceedings{manas2024tr2mtl,
  title={TR2MTL: LLM based framework for metric temporal logic formalization of traffic rules},
  author={Manas, Kumar and Zwicklbauer, Stefan and Paschke, Adrian},
  booktitle={2024 IEEE Intelligent Vehicles Symposium (IV)},
  pages={1206--1213},
  year={2024},
  organization={IEEE}
}

@inproceedings{zin2025towards,
  title={Towards Machine-Readable Traffic Laws: Formalizing Traffic Rules into PROLOG Using LLMs},
  author={Zin, May Myo and Borges, Georg and Satoh, Ken and Fungwacharakorn, Wachara},
  booktitle={Proceedings of the Twentieth International Conference on Artificial Intelligence and Law},
  pages={327--336},
  year={2025}
}

@incollection{fuchs2008attempto,
  title={Attempto controlled english for knowledge representation},
  author={Fuchs, Norbert E and Kaljurand, Kaarel and Kuhn, Tobias},
  booktitle={Reasoning Web: 4th International Summer School 2008, Venice, Italy, September 7-11, 2008, Tutorial Lectures},
  pages={104--124},
  year={2008},
  publisher={Springer}
}

@incollection{kowalski2023logical,
  title={Logical English for law and education},
  author={Kowalski, Robert and D{\'a}vila, Jacinto and Sartor, Galileo and Calejo, Miguel},
  booktitle={Prolog: The Next 50 Years},
  pages={287--299},
  year={2023},
  publisher={Springer}
}

@inproceedings{rizaldi2015formalising,
  title={Formalising traffic rules for accountability of autonomous vehicles},
  author={Rizaldi, Albert and Althoff, Matthias},
  booktitle={2015 IEEE 18th international conference on intelligent transportation systems},
  pages={1658--1665},
  year={2015},
  organization={IEEE}
}

@inproceedings{rizaldi2017formalising,
  title={Formalising and monitoring traffic rules for autonomous vehicles in Isabelle/HOL},
  author={Rizaldi, Albert and Keinholz, Jonas and Huber, Monika and Feldle, Jochen and Immler, Fabian and Althoff, Matthias and Hilgendorf, Eric and Nipkow, Tobias},
  booktitle={International conference on integrated formal methods},
  pages={50--66},
  year={2017},
  organization={Springer}
}

@inproceedings{linker2017spatial,
  title={Spatial reasoning about motorway traffic safety with Isabelle/HOL},
  author={Linker, Sven},
  booktitle={International Conference on Integrated Formal Methods},
  pages={34--49},
  year={2017},
  organization={Springer}
}

@inproceedings{esterle2020formalizing,
  title={Formalizing traffic rules for machine interpretability},
  author={Esterle, Klemens and Gressenbuch, Luis and Knoll, Alois},
  booktitle={2020 IEEE 3rd Connected and Automated Vehicles Symposium (CAVS)},
  pages={1--7},
  year={2020},
  organization={IEEE}
}

@inproceedings{hilscher2011abstract,
  title={An abstract model for proving safety of multi-lane traffic manoeuvres},
  author={Hilscher, Martin and Linker, Sven and Olderog, Ernst-R{\"u}diger and Ravn, Anders P},
  booktitle={International Conference on Formal Engineering Methods},
  pages={404--419},
  year={2011},
  organization={Springer}
}

@article{shalev2017formal,
  title={On a formal model of safe and scalable self-driving cars},
  author={Shalev-Shwartz, Shai and Shammah, Shaked and Shashua, Amnon},
  journal={arXiv preprint arXiv:1708.06374},
  year={2017}
}

@inproceedings{censi2019liability,
  title={Liability, ethics, and culture-aware behavior specification using rulebooks},
  author={Censi, Andrea and Slutsky, Konstantin and Wongpiromsarn, Tichakorn and Yershov, Dmitry and Pendleton, Scott and Fu, James and Frazzoli, Emilio},
  booktitle={2019 international conference on robotics and automation (ICRA)},
  pages={8536--8542},
  year={2019},
  organization={IEEE}
}

@inproceedings{maierhofer2020formalization,
  title={Formalization of interstate traffic rules in temporal logic},
  author={Maierhofer, Sebastian and Rettinger, Anna-Katharina and Mayer, Eva Charlotte and Althoff, Matthias},
  booktitle={2020 IEEE Intelligent Vehicles Symposium (IV)},
  pages={752--759},
  year={2020},
  organization={IEEE}
}

@article{wan2024semantic,
  title={Semantic consistency and correctness verification of digital traffic rules},
  author={Wan, Lei and Wang, Changjun and Luo, Daxin and Liu, Hang and Ma, Sha and Hu, Weichao},
  journal={Engineering},
  volume={33},
  pages={47--62},
  year={2024},
  publisher={Elsevier}
}

@inproceedings{palmirani2011legalruleml,
  title={Legalruleml: Xml-based rules and norms},
  author={Palmirani, Monica and Governatori, Guido and Rotolo, Antonino and Tabet, Said and Boley, Harold and Paschke, Adrian},
  booktitle={International Workshop on Rules and Rule Markup Languages for the Semantic Web},
  pages={298--312},
  year={2011},
  organization={Springer}
}

@inproceedings{ferraro2019automatic,
  title={Automatic extraction of legal norms: Evaluation of natural language processing tools},
  author={Ferraro, Gabriela and Lam, Ho-Pun and Tosatto, Silvano Colombo and Olivieri, Francesco and Islam, Mohammad Badiul and van Beest, Nick and Governatori, Guido},
  booktitle={JSAI International Symposium on Artificial Intelligence},
  pages={64--81},
  year={2019},
  organization={Springer}
}

@article{englishgrammar,
  title={Grammar-forced translation of natural language to temporal logic using LLMs},
  author={English, William and Simon, Dominic and Jha, Sumit Kumar and Ewetz, Rickard},
  journal={arXiv preprint arXiv:2512.16814},
  year={2025}
}
\end{document}